\documentclass[10pt,journal,compsoc]{IEEEtran}

\usepackage{graphicx}
\usepackage{comment}
\usepackage{epsfig}
\usepackage{amsmath,amssymb} 
\usepackage{color}
\usepackage[table]{xcolor}
\usepackage{tikz}
\usepackage{comment}
\usepackage{multirow}
\usepackage[hang,flushmargin]{footmisc} 
\usepackage{booktabs,tabularx}
\usepackage{enumitem}
\usepackage{makecell}
\usepackage{algorithm}
\usepackage{algorithmic}
\usepackage{threeparttable}
\usepackage{lipsum}
\usepackage[pagebackref=false, breaklinks=true, colorlinks, bookmarks=false]{hyperref}
\usepackage{url}

\usepackage{graphicx}
\usepackage{amsmath}
\usepackage{amssymb}
\usepackage{booktabs}
\usepackage{multirow}
\usepackage{stfloats}
\usepackage{caption}
\usepackage{colortbl}
\usepackage{multirow}
\usepackage{multicol}
 \usepackage{subcaption} 
 \usepackage{pifont}
\usepackage{marvosym}
\usepackage{tablefootnote}

\definecolor{linecolor}{rgb}{0.82, 0.94, 0.75}

\usepackage{booktabs} 

\definecolor{greenbg}{rgb}{0.9, 1.0, 0.9} 
\definecolor{cvprblue}{rgb}{0.21,0.49,0.74}
\definecolor{kaiming-green}{RGB}{57,181,74} 

\definecolor{linecolor}{RGB}{224, 224, 224}

\definecolor{mamba}{RGB}{153, 151, 239}
\definecolor{pretty-blue}{RGB}{0, 113, 188}

\def\ours{{EasyCache}}

\definecolor{lowred}{RGB}{238,18,137}

\definecolor{lowerred}{RGB}{255,110,180}

\definecolor{rowunit}{RGB}{128,128,255}
\definecolor{evaunit01green}{RGB}{54,125,189}

\makeatletter
\renewcommand{\maketag@@@}[1]{\hbox{\m@th\normalsize\normalfont#1}}%
\makeatother

\ifCLASSOPTIONcompsoc
  \usepackage[nocompress]{cite}
\else
  \usepackage{cite}
\fi

\ifCLASSINFOpdf
\else
\fi

\begin{document}

\title{
Less is Enough: Training-Free Video Diffusion Acceleration via Runtime-Adaptive Caching
}

\author{
Xin Zhou, Dingkang Liang, Kaijin Chen, Tianrui Feng, Xiwu Chen, Hongkai Lin, Yikang Ding, \\ Feiyang Tan, Hengshuang Zhao, Xiang Bai, \textit{Senior Member, IEEE}
\IEEEcompsocitemizethanks{
\IEEEcompsocthanksitem X. Zhou, D. Liang, K. Chen, T. Feng, H. Lin, and X. Bai are with Huazhong University of Science and Technology. 

E-mail: (xzhou03, dkliang, xbai)@hust.edu.cn

\IEEEcompsocthanksitem X. Chen, Y. Ding, and F. Tan are with MEGVII Technology.

\IEEEcompsocthanksitem H. Zhao is with the University of Hong Kong.

\IEEEcompsocthanksitem Xin Zhou and Dingkang Liang make equal contributions. \\
The corresponding author is Xiang Bai (xbai@hust.edu.cn).

}
}

\markboth{
}%
{Shell \MakeLowercase{\textit{et al.}}: Bare Advanced Demo of IEEEtran.cls for IEEE Computer Society Journals}

\IEEEtitleabstractindextext{%
\begin{abstract}

Video generation models have demonstrated remarkable performance, yet their broader adoption remains constrained by slow inference speeds and substantial computational costs, primarily due to the iterative nature of the denoising process. Addressing this bottleneck is essential for democratizing advanced video synthesis technologies and enabling their integration into real-world applications. This work proposes \ours, a training-free acceleration framework for video diffusion models. \ours~introduces a lightweight, runtime-adaptive caching mechanism that dynamically reuses previously computed transformation vectors, avoiding redundant computations during inference. Unlike prior approaches, \ours~requires no offline profiling, pre-computation, or extensive parameter tuning. We conduct comprehensive studies on various large-scale video generation models, including OpenSora, Wan2.1, and HunyuanVideo. Our method achieves leading acceleration performance, reducing inference time by up to 2.1-3.3$\times$ compared to the original baselines while maintaining high visual fidelity with a significant up to 36\% PSNR improvement compared to the previous SOTA method. This improvement makes our \ours~a efficient and highly accessible solution for high-quality video generation in both research and practical applications. The code is available at \url{https://github.com/H-EmbodVis/EasyCache}.

\end{abstract}

}

\maketitle

\IEEEdisplaynontitleabstractindextext

%
\IEEEpeerreviewmaketitle

\ifCLASSOPTIONcompsoc
\IEEEraisesectionheading{\section{Introduction}\label{sec:introduction}}
\else
\section{Introduction}
\label{sec:introduction}
\fi


%
%
%
%

\begin{figure}[!t]
	\begin{center}
		\includegraphics[width=1.0\linewidth]{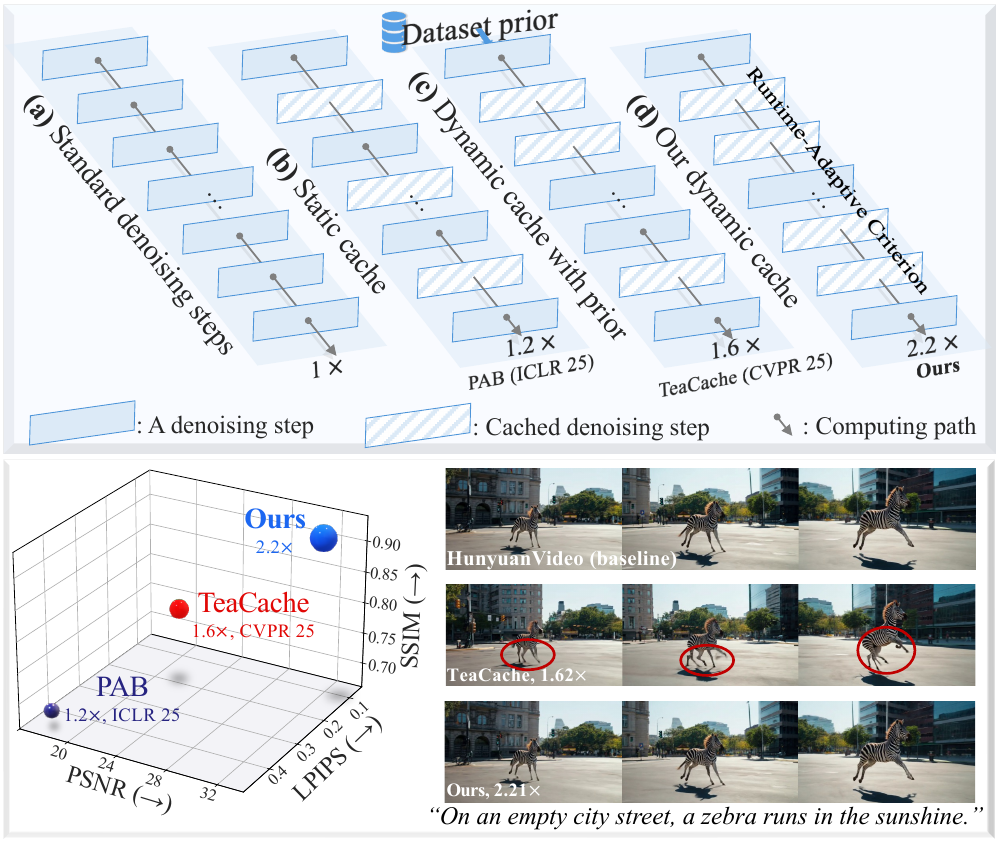}
	\end{center}
        \vspace{-10pt}
	\caption{The comparison between (a) the default iterative denoising, (b) static caching with fixed intervals, (c) dynamic cache with external ``Dataset prior", and (d) our dynamic cache reuses computation by a runtime-adaptive criterion.}
	\label{fig:intro}
    \vspace{-10pt}
\end{figure}

\IEEEPARstart{V}{ideo} generation~\cite{zheng2024open,ma2024latte,brooks2024video,wan2025} has become a central task in generative modeling, with applications in digital content creation and virtual environment interaction. Recently, Diffusion Transformers (DiTs)\cite{peebles2023scalable} have emerged as the dominant paradigm in this domain, credited for their remarkable representational capability and scalability. Nevertheless, the practical adoption of DiTs faces significant hurdles due to high computational costs and prolonged inference durations, primarily resulting from the iterative denoising steps required to model complex spatiotemporal interactions (as in Fig.\ref{fig:intro}(a)). For example, generating a 5s, 720P video with HunyuanVideo~\cite{kong2024hunyuanvideo} requires $\sim$2 H20 GPU hours. These computational demands substantially limit their feasibility in real-time scenarios and resource-constrained environments, thus restricting broader industrial and practical utilization.

The community has explored various acceleration approaches~\cite{lu2022dpm,korthikanti2023reducing,meng2023distillation,zhang2025sageattention} to mitigate computational barriers. While methods based on distillation or architectural changes can accelerate inference, they require extensive training datasets and significant resources. This motivates growing interest in training-free techniques. Among these, feature caching, which reuses intermediate results across denoising steps, has shown considerable promise. Some prior works~\cite{xu2018deepcache,zhao2024real,chen2024delta,selvaraju2024fora} relied on static caching schemes, reusing computations at fixed intervals (as depicted in Fig.~\ref{fig:intro}(b)). However, such uniform caching lacks adaptability and often fails to align with the dynamic behavior of generative models, limiting cache effectiveness when output variations differ significantly across timesteps.

A recent pioneer work, TeaCache~\cite{liu2025timestep}, shifts the focus to dynamic caching strategies, which leverage polynomial fitting and extensive offline profiling of dataset-specific priors to estimate suitable cache intervals (Fig.~\ref{fig:intro}(c)). Although flexible, TeaCache suffers from practical limitations. Its reliance on extensive offline profiling and meticulous hyperparameter tuning makes it sensitive to dataset distributions, undermining its generalizability. Moreover, the dependency on the dataset prior may misalign caching strategies with the model's internal dynamics, compromising generation fidelity, especially with aggressive feature reuse. These challenges naturally lead to a critical research question: \textit{How can we design a runtime-adaptive caching framework that accelerates DiT-based video generation while preserving generation fidelity?}

To answer this question, we begin by analyzing the internal feature dynamics of DiTs during the iterative generation process, revealing a notable stability in model transformation rates (the ratio of the change in predicted noise and latent input) throughout denoising. Inspired by this stability, we posit that intermediate outputs can be reliably estimated using previously computed transformations. This motivates the design of our \textbf{\ours}, a simple yet effective feature caching framework that enables adaptive computation reuse for accelerating inference without any extra training. 

Specifically, \ours~monitors the relative transformation rate and detects local stability via a dynamic, runtime-adaptive thresholding scheme. When stability is estimated, it reuses previously computed transformation vectors to approximate future outputs, thus avoiding redundant full-model evaluations. Crucially, this approximation is governed by an accumulated local error indicator, which ensures that computation reuse is only triggered under reliable conditions. Unlike prior methods, which rely on costly offline statistics and fixed heuristics (e.g., polynomial fitting), \ours~makes fully online decisions based on lightweight first-order dynamics, requiring no retraining, dataset-specific tuning, and architectural modification.

Extensive experiments on several video generation models demonstrate that \ours~achieves significant speedups and outstanding visual retention with original videos. On the computationally demanding HunyuanVideo~\cite{kong2024hunyuanvideo}, \ours~improves the speedup over the SOTA caching strategy TeaCache~\cite{liu2025timestep} by 36\% while boosting the PSNR by 37\% and SSIM by 14\%. Moreover, \ours~is able to achieve continuous acceleration on the already accelerated efficient attention~\cite{xi2025sparse}, pushing the speedup to 3.3$\times$ with only a slight 0.3 PSNR drop.

Our main contributions are summarized as follows: \textbf{1)} We reveal an exploitable relative stability in the transformation rates throughout the denoising process for Diffusion Transformers. \textbf{2)} We propose \ours, a training-free feature caching framework that exploits this stability through a lightweight, runtime-adaptive criterion, enabling substantial inference acceleration with minimal compromising model integrity. \textbf{3)} Extensive experiments across various video generation models demonstrate that our \ours~achieves significant speedups while faithfully preserving visual fidelity, outperforming SOTA caching methods in both inference speed and visual retention. Moreover, \ours~is orthogonally compatible with other acceleration strategies, further enhancing its practical applicability.

\section{Related work}
\subsection{Diffusion Model}
Diffusion models~\cite{ho2020denoising,peebles2023scalable} have revolutionized generative modeling across diverse domains, becoming a dominant paradigm for high-quality content synthesis. These probabilistic models have achieved remarkable success in fields including image generation~\cite{saharia2022photorealistic,rombach2022high,podell2023sdxl}, 3D object synthesis~\cite{zhao2025hunyuan3d,li2025step1x,chen20253dtopia}, and video creation~\cite{zhou2022magicvideo,blattmann2023align,blattmann2023stable,hong2023cogvideo}.

Within this landscape, video generation has garnered significant attention due to the growing demand for dynamic content creation. However, generating coherent and high-fidelity video presents significant challenges due to the complex temporal dynamics and the substantial computational overhead with long sequences. As a result, the field has witnessed a notable shift from traditional U-Net towards more scalable Diffusion Transformers (DiTs)~\cite{peebles2023scalable}, which have demonstrated strong capabilities in handling complex data modalities. DiT-based video generative models like Sora~\cite{brooks2024video}, HunyuanVideo~\cite{kong2024hunyuanvideo}, and Wan2.1~\cite{wan2025} have demonstrated exceptional capabilities in modeling complex spatiotemporal relationships. Despite their power, the iterative nature of these models leads to significant inference latency and remains a key adoption barrier, necessitating further research into efficient generation methods.

\subsection{Training-free Diffusion Inference Acceleration}
Training-free inference acceleration techniques~\cite{songdenoising,karras2022elucidating, lu2022dpm,bolya2023token, wang2024attention, zhang2025training,zou2024accelerating,zhang2025sageattention} are achieving significant attention as they speed up inference without resource-intensive retraining, thus preserving the integrity of large-scale pre-trained models. Many methods are designed to reduce per-step costs. For example, efficient attention implementation like SVG~\cite{xi2025sparse} utilizes inherent sparsity in the 3D full attention, training-freely achieving promising inference speed amelioration.

Another representative approach for reducing per-step cost is feature caching, which exploits computational redundancy in the iterative denoising process. Early static caching~\cite{selvaraju2024fora,chen2024delta} uses pre-determined rules but lacks the adaptability to non-uniform process dynamics. For example, PAB~\cite{zhao2024real} employs a pyramid-style heuristic to broadcast attention at fixed intervals. Dynamic caching~\cite{wimbauer2024cache,liu2025timestep} is then proposed to address this rigidity. The current SOTA, TeaCache~\cite{liu2025timestep}, exemplifies a dynamic strategy that relies on an external dataset prior, modeling the relationship between timestep embeddings and output changes with a pre-fitted polynomial. While more adaptive, this reliance on offline profiling hinders its generalizability. Our work is motivated to develop a fully dynamic caching framework that eliminates the need for external priors by leveraging the intrinsic runtime dynamics of the inference process.

\section{\ours}
This section introduces our training-free runtime-adaptive dynamic caching framework for Diffusion Transformer (DiT)-based video generation.

\subsection{A Revisit of Diffusion Modeling with Transformer}

The Diffusion Transformer (DiT)~\cite{peebles2023scalable} integrates Transformer architectures into diffusion models~\cite{ho2020denoising,song2021score}, achieving superior scalability and performance for video generation. In this paradigm, noisy latent input $\mathbf{x}_t$ at $t$-th step is partitioned into non-overlapping patches, processed by sequential Transformer layers, and modulated by a learned time embedding $\mathbf{e}_t$ that reflects the noise level. During inference, a latent variable sampled from Gaussian noise is iteratively denoised over $T$ steps to produce the final data sample. However, DiT-based video generation remains computationally intensive due to the need to model complex spatiotemporal dependencies and perform iterative denoising over many steps. For instance, generating a 5s, 720P video with HunyuanVideo~\cite{kong2024hunyuanvideo} or Wan2.1-14B~\cite{wan2025} requires $\sim$2 H20 hours. Such high computational demands significantly impede applications requiring user interaction or low latency. Although model-specific fine-tuning or retraining can improve efficiency, these methods demand substantial computational resources and large-scale datasets, limiting their accessibility. Therefore, it is critical to develop efficient, training-free acceleration techniques that can reduce inference costs while preserving generation quality.

\subsection{Investigation of Feature Dynamics} \label{sec:investigation}
\begin{figure}[!t]
	\begin{center}
		\includegraphics[width=\linewidth]{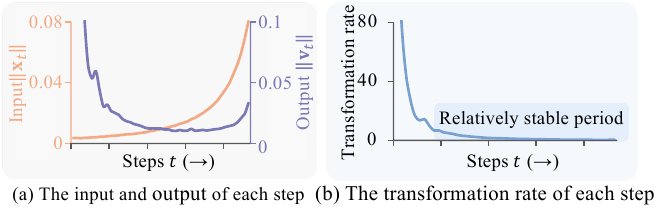}
	\end{center}
        \vspace{-10pt}
	\caption{Analysis of feature dynamics. (a) The L1 norm of the input and output of each step. (b) The changes in the relative transformation rate between consecutive steps.}
    \vspace{-10pt}
	\label{fig:investigate}
\end{figure}
To develop a reliable protocol for dynamic feature caching, we first systematically analyze the feature dynamics within diffusion transformers (DiTs). During inference, the network $u_{\theta}$ estimates the noise component $\mathbf{v}_t$ from the noisy latent $\mathbf{x}_t$ at each step $t$, conditioned on a prompt $\mathcal{T}$. The iteration at step $t$ can be simply referred to $\mathbf{v}_t = u_{\theta}(\mathbf{x}_t \mid \mathcal{T})$.

\textbf{Feature Dynamics Investigation.} 
The most straightforward metric to investigate the correlation of DiT's internal feature is to evaluate feature dynamics as the norm of the input and output for each step, i.e., $\|\mathbf{x}_t\|$ and $\|\mathbf{v}_t\|$, where $\|\cdot\|$ denotes the L1 norm and an average operation. As shown in Fig.~\ref{fig:investigate}(a), $\|\mathbf{x}_t\|$ and $\|\mathbf{v}_t\|$ exhibit a growth trend and U-shaped curve over the diffusion process, respectively, showing a completely different changing pattern and making a difficult feature dynamic estimation.

To mitigate this issue, TeaCache~\cite{liu2025timestep} estimates output dynamics for feature caching using polynomial fitting to determine scaling factors between input and output relative differences. However, this approach requires heuristic factor fitting and separate fitting for different settings (e.g., resolution changes).

\textbf{Transformation Rate Stability.} \label{sec:rate-stability}
To establish a criterion for adaptive computation reuse, we quantify the stability of the DiT's output with respect to input changes. Inspired by the concept of function derivatives, we approximate the ``\textit{directional derivative}" of the model $u_{\theta}$ with respect to input latent $\mathbf{x}_t$ using a first-order approximation:
\begin{equation}\label{eq:transformation-derivative}
\frac{\partial u_{\theta}}{\partial \mathbf{x}_t}\approx \frac{u_{\theta}(\mathbf{x}_t \mid \mathcal{T}) - u_{\theta}(\mathbf{x}_{t-1} \mid \mathcal{T})}{\mathbf{x}_t - \mathbf{x}_{t-1}} = \frac{\mathbf{v}_t - \mathbf{v}_{t-1}}{\mathbf{x}_t - \mathbf{x}_{t-1}}.
\end{equation}
This represents the ratio of the output change to the input change between consecutive steps. To better access and simplify this ``derivative", we adopt the L1 norm and average to reduce the derivative at step $t$ to a number by:
\begin{equation}\label{eq:transformation-rate}
k_t = \frac{\|\mathbf{v}_t - \mathbf{v}_{t-1}\|}{\|\mathbf{x}_t - \mathbf{x}_{t-1}\|}.
\end{equation}
While $k_t$ is not a formal derivative, it serves a similar purpose, and we define it as the relative transformation rate.

Surprisingly, as shown in Fig.~\ref{fig:investigate}(b), $k_t$ rapidly stabilizes and remains relatively stable after an initial period of sharp fluctuation. We hypothesize that this reflects the DiT model's evolving role: early steps focus on establishing the global layout with a non-linear mapping, while later steps refine local details with a stable, approximately linear relationship. This largely relative linear relationship across a wide range of the denoising process motivates our method, suggesting reliable output estimation and enabling runtime-adaptive criterion of redundant computations.

\begin{figure*}[!t]
	\begin{center}
		\includegraphics[width=0.98\linewidth]{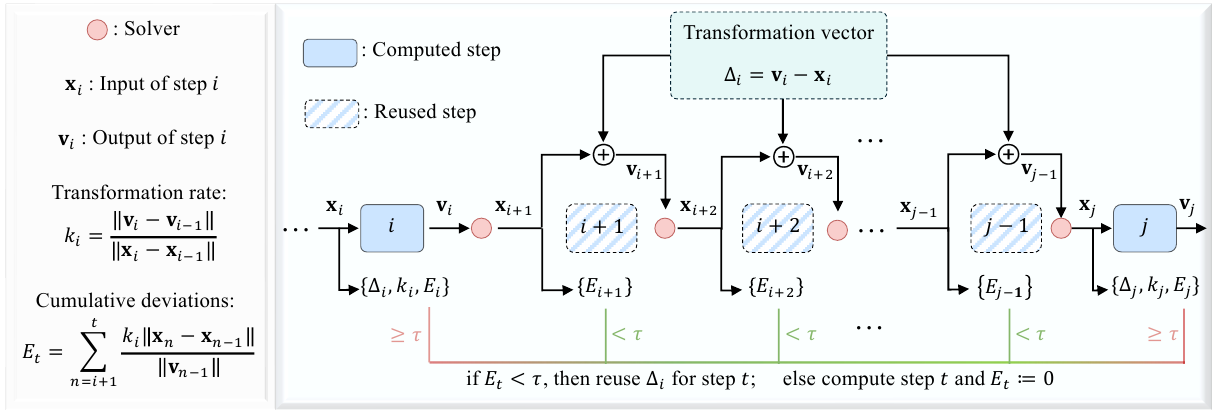}
	\end{center}
        \vspace{-10pt}
        \caption{The overall of our method. For simplicity, we start from a computed step $i$. A runtime-adaptive criterion evaluates each subsequent step, reusing the cached transformation vector $\Delta_i$ while the accumulated deviation $E_t$ remains below a threshold $\tau$. A full computation is performed when the threshold is exceeded, as exemplified in step $j$.}
        \vspace{-10pt}
	\label{fig:pipeline}
\end{figure*}
\subsection{Runtime-Adaptive Criterion} \label{sec:ours}
Capitalizing on the observed stability of the transformation rate, we propel forward with introducing a simple yet effective adaptive computation reuse framework, denoted as \ours. It operates on a core principle, i.e., approximating the model's output using previously computed transformations during stable diffusion phases, employing a lightweight criterion to ensure approximation accuracy. 

\textbf{Efficient Output Approximation.} 
The foundation of \ours~lies in the observation from Sec.~\ref{sec:rate-stability} that the transformation of $\mathbf{x}_t$ into predicted noise $\mathbf{v}_t$ is expected to exhibit local consistency during a relatively stable phase. We define the transformation vector at step $t$ as $\Delta_t := \mathbf{v}_t - \mathbf{x}_t$. 

Our central hypothesis is that in the relatively stable regions of the diffusion process, the current transformation vector $\Delta_t$ can be well approximated by one from a recent fully computed step $i < t$ by $\Delta_t \approx \Delta_{i}$.
Thus, for steps deemed skippable, we approximate the model output $\mathbf{v}_t$ using this cached transformation vector $\Delta_{i}$:
\begin{equation}
    \hat{\mathbf{v}}_t = \mathbf{x}_t + \Delta_{i}, \quad \text{where } \Delta_{i} := \mathbf{v}_{i} - \mathbf{x}_{i}.
    \label{eq:output_approx}
\end{equation}
This approximation incurs negligible computational cost compared to a full forward pass $u_\theta(\mathbf{x}_t | \mathcal{T})$. The critical challenge, addressed next, is to establish a lightweight yet reliable criterion for determining when this reuse is appropriate without sacrificing generation quality.

\textbf{Adaptive Caching Criterion.}
The approximation in Eq.~\ref{eq:output_approx} relies on an adaptive criterion to determine whether the transformation vector remains nearly unchanged, i.e., whether $|| \Delta_t - \Delta_{i} ||$  $(t>i\ge1)$ is small. However, directly calculating this difference at every step is non-trivial. To address this, we introduce a proxy measure informed by the inherent dynamics of the diffusion process.

As shown in Fig.~\ref{fig:pipeline}, our goal is to propose an adaptive caching criterion that governs feature reuse for any step $t$ following the most recent fully computed step, denoted as $i$. Our central hypothesis is that if the expected change in the model output $\mathbf{v}_t$ relative to $\mathbf{v}_{t-1}$ is small, the process is considered locally relatively stable. This stability, in turn, implies that the transformation vector $\Delta_t$ also remains nearly constant, thus justifying the safe reuse of the cached transformation vector $\Delta_i$ from step $i$.

To estimate the output for step $t$, we approximate $||\mathbf{v}_t - \mathbf{v}_{t-1}||$ as $k_{i} ||\mathbf{x}_t - \mathbf{x}_{t-1}||$ by making a constant assumption for $k_{i}$. Based on this, we define the local stability indicator (output change rate) $\varepsilon_t$ as the percentage of the estimated output change at step $t$ relative to the latest output $\mathbf{v}_{t-1}$:
\begin{equation}
    \varepsilon_t = \frac{||\mathbf{v}_t - \mathbf{v}_{t-1}||}{||\mathbf{v}_{t-1}||}\approx \frac{k_{i} ||\mathbf{x}_t - \mathbf{x}_{t-1}||}{||\mathbf{v}_{t-1}||}\times 100\%.
    \label{eq:epsilon_t}
\end{equation}
A small $\varepsilon_t$ indicates that the output is evolving smoothly and predictably with respect to input variations. To account for cumulative deviations since $i$, we sum these indicators:
\begin{equation}
    E_t = \sum_{n=i+1}^{t} \varepsilon_n.
    \label{eq:E_t}    
\end{equation}
The transformation $\Delta_{i}$ is reused for step $t$ if this accumulated stability indicator $E_t$ remains below a predefined tolerance threshold $\tau$. Thus, $E_t < \tau$ is the practical condition for deeming the process stable enough for approximation. A larger $\tau$ allows for greater tolerance to cumulative deviations, enabling more aggressive feature reuse, while a smaller $\tau$ potentially promotes visual fidelity through conservative reuse. As a key hyperparameter, $\tau$ can be easily tuned for a trade-off between speed and quality.

\textbf{The Integrated \ours~Framework.}
Finally, the complete \ours~integrates the runtime-adaptive criterion and the approximation strategy into a unified workflow. The update rule for $\mathbf{v}_t$ at any given step $t$ is as follows:
\begin{equation} \label{eq:easycache_update_rule}
\hspace{-3.5mm}
\resizebox{.9\hsize}{!}{$\mathbf{v}_{t} = 
\begin{cases} 
u_{\theta}(\mathbf{x}_{t}|\mathcal{T}), & \text{if } E_{t} \ge \tau \text{ or } t \in [0, R-1]\cup \{T-1\} \\
\mathbf{x}_{t} + \Delta_{i}, & \text{otherwise}
\end{cases}.$}
\end{equation}
Here, $R$ is the number of initial warm-up steps where full computation is mandatory to capture the highly changing initial diffusion phase. For any step $t$, if $E_{t}\ge \tau$, $t<R$ or $t=T-1$, a full computation $u_{\theta}(\mathbf{x}_{t}|\mathcal{T})$ is performed. Subsequently, the cache is updated by storing the new transformation vector $\Delta_t$, the reference step is set to $i := t$, and the cumulative deviation is reset to $E_t := 0$. Otherwise, the cached vector $\Delta_i$ is reused. Our \ours~ensures that computational resources are dynamically allocated, focusing on steps that introduce significant changes.

\subsection{An Intuitive Perspective on the Relative Stability} \label{sec:theor_analysis}
We analyze the stability of the transformation rate $k$ (Eq.~\ref{eq:transformation-rate}) during video generation, using widely used Ordinary Differential Equation (ODE)-based sampler in video generation field, i.e., flow matching~\cite{lipmanflow}. Typically, the current $\mathbf{x}_t$ is updated via an Euler step: $\mathbf{x}_t = \mathbf{x}_{t-1} + \mathbf{v}_{t-1} \Delta s_t$, where $\Delta s_t$ represents the continuous time step from step $t-1$ to $t$, and $\mathbf{v}_{t-1}$ is the predicted velocity field. Therefore, substituting this into Eq.~\ref{eq:transformation-rate}, we can obtain:
\begin{equation}
    k_t = \frac{\|\mathbf{v}_t - \mathbf{v}_{t-1}\|}{\|\mathbf{x}_{t-1} + \mathbf{v}_{t-1}\Delta s_t - \mathbf{x}_{t-1}\|} = \frac{\|\mathbf{v}_t - \mathbf{v}_{t-1}\|}{\|\mathbf{v}_{t-1} \Delta s_t\|}.
\end{equation}

The stability can be understood as follows: 1) The training process encourages a linear velocity field that smoothly transforms noise into target data. In the later stages of generation, the transformation from $\mathbf{v}_{t-1}$ to $\mathbf{v}_t$ becomes near-linear, approximating $\mathbf{v}_t \approx c  \mathbf{v}_{t-1}$, where $c$ is a scalar or a near-scalar transformation that varies slowly during the stable phase. Thus, $k_t \approx \frac{|c-1|}{\Delta s_t}$; 2) Given that $\Delta s_t$ is typically determined by a fixed scheduler (potentially constant or slowly varying), $k_t$ quickly stabilizes after an initial period of fluctuation during non-linear structure formation. This relative stability enables the training-free, runtime-adaptive identification of redundant computation steps, allowing selective reuse of transformation vectors with minimal output sacrifice. As a result, our \ours~substantially reduces inference cost while maintaining strong visual fidelity since approximations are applied when the model’s behavior is intrinsically predictable.

\begin{table*}[!t]
    \centering
    \footnotesize
    \caption{The comparison between other acceleration strategies and our \ours. We report inference latency, speedup, and visual quality metrics (PSNR, SSIM, LPIPS, and VBench) on representative video generation models.}
    \vspace{-7pt}
    \setlength{\tabcolsep}{3.7mm}
    \label{tab:main}
    \begin{tabular}{lccccccc}
        \toprule
        \multirow{2.3}{*}{Methods} & \multirow{2.3}{*}{Reference} & \multicolumn{2}{c}{Efficiency} & \multicolumn{3}{c}{Visual Retention}& \multirow{2.3}{*}{VBench (\%) $\uparrow$} \\ 
        \cmidrule(lr){3-4} \cmidrule(lr){5-7}& & Latency (s) $\downarrow$ & Speedup $\uparrow$ & PSNR $\uparrow$ & SSIM $\uparrow$ & LPIPS $\downarrow$ &  \\
        \midrule
        \multicolumn{8}{c}{Open-Sora 1.2~\cite{zheng2024open} (51 frames, 848$\times$480)} \\
        \midrule
        Open-Sora 1.2 $(T = 30)$& - & 44.90 & 1$\times$ & - & - & - & \textbf{79.40} \\
        + 50\% steps& - & 22.71 & 1.98$\times$ & 15.82 & 0.6961 & 0.3363 & 77.36 \\
        + Random 0.5& - & 22.68 & 1.98$\times$ & 16.51 & 0.7037 & 0.3264 & 76.78 \\
        + Static cache& - & 24.19 & 1.86$\times$ & 15.73 & 0.6961 & 0.3382 & 77.37 \\
        + T-GATE~\cite{liu2025faster} & TMLR 25 & 39.70 & 1.13$\times$ & 19.55 & 0.6927 & 0.2612 & 75.42 \\
        + PAB~\cite{zhao2024real}& ICLR 25 & 31.57 & 1.42$\times$ & 23.58 & 0.8220 & 0.1743 & 76.95 \\
        + TeaCache~\cite{liu2025timestep}& CVPR 25 & 28.92 & 1.55$\times$ & 23.56 & 0.8433 & 0.1318 & 79.27 \\
        \rowcolor{linecolor}+ \textbf{\ours~(ours)}& - & \textbf{21.21} & \textbf{2.12}$\times$ & \textbf{23.95} & \textbf{0.8556} & \textbf{0.1235} & 78.74 \\
        
        \midrule
        \multicolumn{8}{c}{Wan2.1-1.3B~\cite{wang2025wan} (81 frames, 832$\times$480)} \\
        \midrule
        Wan2.1 $(T = 50)$& - & 175.35 & 1$\times$ & - & - & - & \textbf{81.30} \\
        + 40\% steps& - & 70.10 & 2.50$\times$ & 14.50 & 0.5226 & 0.4374 & 80.30 \\
        + Random 0.4& - & 71.69 & 2.45$\times$ & 11.92 & 0.4204 & 0.5911 & 78.68 \\
        + Static cache& - & 71.54 & 2.45$\times$ & 14.18 & 0.5007 & 0.4789 & 79.58 \\
        + PAB~\cite{zhao2024real}& ICLR 25 & 102.03 & 1.72$\times$ & 18.84 & 0.6484 & 0.3010 & 77.60 \\
        + TeaCache~\cite{liu2025timestep}& CVPR 25 & 87.77 & 2.00$\times$ & 22.57 & 0.8057 & 0.1277 & 81.04 \\
        \rowcolor{linecolor}+ \textbf{\ours~(ours)}& - & \textbf{69.11} & \textbf{2.54}$\times$ & \textbf{25.24} & \textbf{0.8337} & \textbf{0.0952} & 80.49 \\
        \midrule
        \multicolumn{8}{c}{HunyuanVideo~\cite{kong2024hunyuanvideo} (129 frames, 960$\times$544)} \\
        \midrule
        HunyuanVideo $(T = 50)$& - & 1124.30 & 1$\times$ & - & - & - & 82.20 \\
        + 50\% steps& - & 566.17 & 1.99$\times$ & 18.79 & 0.7101 & 0.3319 & 81.78 \\
        + Random 0.5& - & 578.83 & 1.94$\times$ & 19.85 & 0.7201 & 0.3214 & 81.04 \\
        + Static cache& - & 573.38 & 1.96$\times$ & 18.74 & 0.7081 & 0.3309 & 81.76 \\
        + PAB~\cite{zhao2024real}& ICLR 25 & 958.23 & 1.17$\times$ & 18.58 & 0.7023 & 0.3827 & 76.98 \\
        + TeaCache~\cite{liu2025timestep}& CVPR 25 & 674.04 & 1.67$\times$ & 23.85 & 0.8185 & 0.1730 & \textbf{82.32} \\
        + SVG~\cite{xi2025sparse} & ICML 25 & 802.70 & 1.40$\times$ & 26.57 & 0.8596 & 0.1368 & 81.97 \\
        \rowcolor{linecolor}+ \textbf{\ours~(ours)}& - & \textbf{507.97} & \textbf{2.21}$\times$ & \textbf{32.66} & \textbf{0.9313} & \textbf{0.0533} & 82.01 \\
        \bottomrule
    \end{tabular}
\end{table*}

\begin{table}[!t]
\footnotesize
\setlength{\tabcolsep}{0.6mm}
\centering
\caption{Generalization to text-to-image generation.}
\vspace{-7pt}
\label{tab:t2i}
\begin{tabular}{ lcccccc }
\toprule
    \multirow{2.3}{*}{Method} & \multicolumn{2}{c}{Efficiency} & \multicolumn{2}{c}{Visual quality} \\ 
    \cmidrule(lr){2-3} \cmidrule(lr){4-5} 
    & Latency (s) $\uparrow$& Speedup $\uparrow$& FID-30k $\downarrow$ & CLIP Score$\uparrow$\\
    \midrule
    FLUX.1-dev~\cite{flux2024} & 25.5 & 1$\times$ & 25.8 & 26.0 \\
    + Teacache~\cite{liu2025timestep} & 7.8 & 3.27$\times$ & 24.5 & \textbf{26.2} \\
    \rowcolor{linecolor}+ \textbf{\ours~(ours)} & \textbf{5.5} & \textbf{4.64}$\times$ & \textbf{23.2} & 26.1 \\
    \bottomrule
\end{tabular}
\end{table}

\section{Experiments}
\subsection{Experiment Setup}

\textbf{Evaluation Metrics.} We evaluate our \ours~on text-to-video generation and its generalization to text-to-image, focusing on inference efficiency (latency and speedup of the DiT modules) and generation quality. For the primary task of text-to-video generation, we use default prompts in VBench~\cite{huang2024VBench} to assess visual retention. Specifically, we measure pixel-level fidelity, structural similarity, and perceptual consistency using PSNR, SSIM~\cite{wang2004image}, and LPIPS~\cite{zhang2018unreasonable} against the original videos. We also report the VBench score for human perceptual alignment. For image generation, we generate 30,000 samples from COCO-2017~\cite{lin2014microsoft} captions, reporting FID-30k for perceptual quality and CLIP Score~\cite{hessel2021clipscore} for semantic alignment.

\textbf{Implementation Details.} Our \ours~only requires a few hyperparameter tuning. To achieve a desirable trade-off between inference efficiency and generation quality, we set ($\tau$, $R$) as (10\%, 5) for Open-Sora 1.2~\cite{zheng2024open}, (5\%, 10) for Wan2.1~\cite{wang2025wan}, and (2.5\%, 5) for HunyuanVideo~\cite{kong2024hunyuanvideo}. For the principal quantitative evaluations reported using the VBench default prompts, we generate 5 video samples with varying seeds per prompt on NVIDIA A800 GPUs. To expedite ablation studies, we generate a single video per prompt with seed 0, using Wan2.1-1.3B as the baseline.

\subsection{Main Results}
We conduct experiments on leading video generation models, including Open-Sora 1.2~\cite{zheng2024open}, Wan2.1-1.3B~\cite{wang2025wan}, and HunyuanVideo~\cite{kong2024hunyuanvideo}, to rigorously evaluate the acceleration and visual retention of \ours.

We first compare \ours~with several naive training-free acceleration strategies, including direct step reduction (e.g., using only 50\% of the denoising steps), probabilistic caching (Random 0.5), and static interval caching. While these straightforward methods are easy to implement and can be tuned to provide a similar speedup as \ours, their lack of adaptability to the evolving dynamics of the diffusion process leads to significant quality degradation. As shown in Tab.~\ref{tab:main}, static cache and direct step reduction result in over 40\% lower PSNR and more than 20\% reduction in SSIM compared to \ours~on HunyuanVideo. This substantial gap shows that static caching fails to accommodate the non-uniform temporal variation inherent in video generation. In contrast, our \ours~employs a runtime-adaptive criterion to selectively reuse computation during locally relatively stable phases in a training-free manner, thereby achieving both superior acceleration and markedly better visual fidelity across all evaluated baselines.

Compared to advanced training-free acceleration methods such as PAB~\cite{zhao2024real} and TeaCache~\cite{liu2025timestep}, our \ours~demonstrates clear advantages in both efficiency and visual quality. As shown in Tab.~\ref{tab:main}, on Wan2.1-1.3B, \ours~achieves a 2.54$\times$ speedup, substantially faster than TeaCache (2.0$\times$) and PAB (1.7$\times$), while maintaining superior visual fidelity with a 11.8\% improvement in PSNR and a 25.5\% reduction in LPIPS relative to the best-competing method TeaCache. On more time-consuming HunyuanVideo, \ours~delivers a significant 36.9\% higher PSNR and a 69.2\% lower LPIPS compared to TeaCache and a notable 6.09dB PSNR and 57.9\% speedup improvement over the leading efficient attention implementation SVG~\cite{xi2025sparse}, resulting in markedly superior perceptual similarity. We observe that the VBench score~\cite{huang2024VBench} exhibits a negligible drop (e.g., -0.19\% for HunyuanVideo). This can be attributed to the evaluation methodology of certain sub-dimensions within VBench, such as its sensitivity to minor structural variations that are not perceptible to human observers. Importantly, all observed VBench deviations remain below 1\%, indicating that the visual retention achieved by \ours~is well within acceptable bounds for practical use. 

Unlike PAB's pyramid-style attention broadcast at fixed intervals and TeaCache's offline heuristics coming from the dataset prior, our \ours~dynamically identifies and exploits locally relatively stable phases. This enables \ours~to achieve higher acceleration with less perceptual quality sacrifice, setting a new standard for training-free diffusion model inference.

\begin{figure*}[!t]
	\begin{center}
		\includegraphics[width=1.0\linewidth]{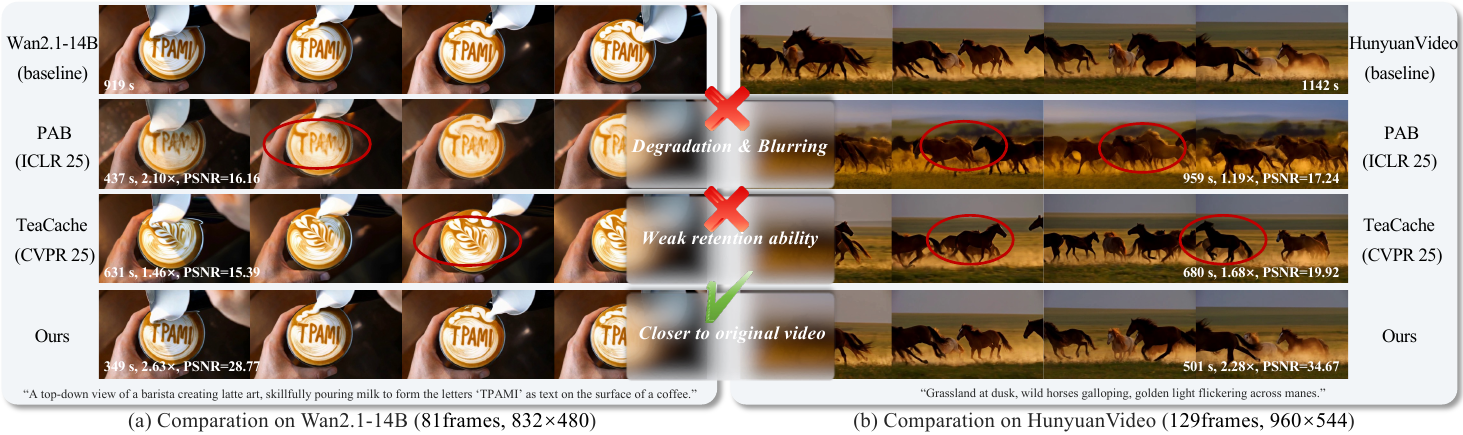}
	\end{center}
        \vspace{-10pt}
	\caption{Qualitative comparison of \ours~with baseline and prior acceleration methods~\cite{zhao2024real,liu2025timestep}. On (a) Wan2.1-14B~\cite{wan2025} and (b) HunyuanVideo~\cite{kong2024hunyuanvideo}. Our approach consistently produces results that are closer to the original video.}
	\label{fig:qualitative_comparison}
        \vspace{-7pt}
\end{figure*}

To further validate the broader applicability of our acceleration strategy, we finally evaluate its performance on text-to-image generation using the 50-step FLUX.1-dev~\cite{flux2024}. As shown in Tab.~\ref{tab:t2i}, \ours~achieves approximately 40\% speedup over TeaCache and markedly reduces latency compared to the FLUX.1-dev baseline. Notably, our approach maintains or even improves generation quality, outperforming FLUX.1-dev by 10\% in FID and 0.1 in CLIP score. These strong performances highlight the intrinsic generalizability of our runtime-adaptive criterion, enabling adequate acceleration and visual quality across both video and image generation.

\subsection{Qualitative Results}

We present qualitative comparisons in Fig.~\ref{fig:qualitative_comparison} with prior caching strategies~\cite{zhao2024real,liu2025timestep} on two computationally intensive baselines, Wan2.1-14B~\cite{wan2025} and HunyuanVideo~\cite{kong2024hunyuanvideo}. Here, we present results for single-sample, noting that such case studies may exhibit more variability than the aggregated metrics in Tab.~\ref{tab:main}. As shown in Fig.~\ref{fig:qualitative_comparison}(a), the less adaptive caching in PAB~\cite{zhao2024real} results in notable degradation and blurring, rendering text and fine details illegible. While dynamic approaches like TeaCache~\cite{liu2025timestep}, which depend on offline dataset priors, frequently fail to retain intricate patterns unique to each sample, as in Fig.~\ref{fig:qualitative_comparison}(b). This limitation emphasizes the difficulty of generalizing to the diverse instance-specific dynamics encountered during inference. In comparison, our \ours, driven by a runtime-adaptive criterion, closely matches the outputs of the original baseline even under aggressive acceleration (e.g., 2.63$\times$ speedup on Wan2.1-14B versus 1.46$\times$ with TeaCache). This superior performance is attributed to leveraging the intrinsic transformation stability of the diffusion process and eliminating the need for offline profiling, thereby ensuring both fidelity and adaptability across models and settings. Additional qualitative results are provided at \url{https://H-EmbodVis.github.io/EasyCache}.

\subsection{Ablation Study}
We conduct ablation studies on the VBench prompts, each generating one video with Wan2.1-1.3B~\cite{wan2025}. Note that the primary objective is to achieve an optimal trade-off between computational efficiency and visual fidelity of the generated videos. The default settings are marked in \colorbox{linecolor}{gray}.

\begin{table}[t]
    \footnotesize
    \caption{Ablation on the error threshold $\tau$.}
    \vspace{-7pt}
    \label{tab:ablation_tau}
    \setlength{\tabcolsep}{1.7mm}
    \centering
    \begin{tabular}{cccccc}
    \toprule
            \multirow{2.3}{*}{$\tau$} & \multicolumn{2}{c}{Efficiency} & \multicolumn{3}{c}{Visual Retention} \\ 
            \cmidrule(lr){2-3} \cmidrule(lr){4-6} 
            & Latency (s) $\downarrow$ & Speedup $\uparrow$ & PSNR $\uparrow$ & SSIM $\uparrow$ & LPIPS $\downarrow$ \\
            \midrule
            2\% & 109.18& 1.61$\times$ & 30.73 & 0.9342 & 0.0356\\
            3\% & 89.51& 1.96$\times$ & 27.58 & 0.8882 & 0.0633\\
            \rowcolor{linecolor}5\% & 69.11 & 2.54$\times$ & 24.74 & 0.8041 & 0.1121\\
            7\% & 63.57 & 2.76$\times$ & 23.22 & 0.7328 & 0.1541\\
            10\% & 56.77 & 3.09$\times$ & 21.67 & 0.6428 & 0.2159\\
            \bottomrule
    \end{tabular}
    \vspace{-10pt}
\end{table}

\textbf{The effect of tolerance threshold $\tau$.}
We first analyze the effect of the tolerance threshold $\tau$, which dictates the aggressiveness of computation reuse in Tab.~\ref{tab:ablation_tau}. A smaller $\tau$ (e.g., 2\%) yields better visual retention (PSNR 30.73) but substantially lower speedup (1.61$\times$). While a higher $\tau=10\%$ significantly boosts speedup with a 21.7\% improvement over $\tau=5\%$, but at the cost of considerable degradation in visual quality. Our default $\tau=5\%$ achieves a compelling trade-off, providing a 2.54$\times$ speedup while maintaining reasonable visual fidelity (PSNR 24.74), effectively balancing efficiency gains with quality preservation.

\textbf{The effect of initial warm-up steps $R$.}
The warm-up steps $R$ controls the number of fully computed initial steps. As discussed in Sec.~\ref{sec:rate-stability}, the initial diffusion phase exhibits rapid and unstable changes in transformation rate changes, and Tab.~\ref{tab:ablation_R} shows that fully computing initial steps is beneficial. Reducing $R$ slightly increases speedup but also causes visual quality degradation. Increasing $R$ to 15 improves visual metrics but reduces the speedup to 2.10$\times$ due to more full computations. We find that $R=10$ in Wan2.1 provides a good balance, ensuring a more stable transformation rate $k$ for reliable subsequent computation reuse, preserving visual quality while achieving a solid 2.54$\times$ speedup.

\textbf{The effect of adaptive reuse criterion.}
We further evaluate different adaptive reuse criteria in Tab.~\ref{tab:ablation_criteria}, using $R=10$ initial steps and similar speedups ($\sim$2.5$\times$) for fair comparison. The probabilistic caching baseline, which reuses computations with a fixed probability, shows suboptimal visual retention, indicating the limitations of static strategies. The output-relative method~\cite{wimbauer2024cache}, which measures reuse based on the relative output change $\|\mathbf{v}_{t-1}-\mathbf{v}_{t-2}\|/\|\mathbf{v}_{t-1}\|$, considers recent output history and achieves moderate performance but is less effective due to the U-shaped change pattern observed in Fig.~\ref{fig:investigate}(a). We also evaluate a without re-computing variant, where the first 20 steps are fully computed, and all subsequent steps directly reuse the transformation vector without any correction. Although the changes in later steps are relatively stable, this still leads to the most degradation in visual quality due to accumulated errors. In contrast, our method dynamically adjusts reuse based on the relationship between input and output changes, effectively correcting potential errors and achieving the best balance between efficiency and fidelity.

\begin{table}[!t]
    \centering
    \footnotesize
    \caption{Ablation on initial warm-up steps number $R$.}
    \vspace{-7pt}
    \label{tab:ablation_R}
    \setlength{\tabcolsep}{2.mm}
    \centering
    \begin{tabular}{cccccc}
    \toprule
            \multirow{2.3}{*}{$R$} & \multicolumn{2}{c}{Efficiency} & \multicolumn{3}{c}{Visual Retention} \\ 
            \cmidrule(lr){2-3} \cmidrule(lr){4-6} 
            & Latency (s) $\downarrow$ & Speedup $\uparrow$ & PSNR $\uparrow$ & SSIM $\uparrow$ & LPIPS $\downarrow$ \\
            \midrule
            2 &  66.73 & 2.63$\times$ & 23.39 & 0.7665 & 0.1389 \\
            5 &  67.21 & 2.61$\times$ & 23.45 & 0.7689 & 0.1374 \\
            \rowcolor{linecolor}10 & 69.11 & 2.54$\times$ & 24.74 & 0.8041 & 0.1121 \\
            15 & 83.49 & 2.10$\times$ & 26.48 & 0.8372 & 0.0874 \\
            \bottomrule
    \end{tabular}
\end{table}

\begin{table}[!t]
    \centering
    \footnotesize
    \caption{Ablation on criteria for adaptive reuse.}
    \vspace{-7pt}
    \label{tab:ablation_criteria}
    \setlength{\tabcolsep}{0.5mm}
    \begin{tabular}{cccccc}
    \toprule
            \multirow{2.3}{*}{\makecell{Adaptive \\ Criterion}} & \multicolumn{2}{c}{Efficiency} & \multicolumn{3}{c}{Visual Retention} \\ 
            \cmidrule(lr){2-3} \cmidrule(lr){4-6} 
            & Latency (s) $\downarrow$ & Speedup $\uparrow$ & PSNR $\uparrow$ & SSIM $\uparrow$ & LPIPS $\downarrow$ \\
            \midrule
            Probabilistic & 70.39 & 2.49$\times$ & 22.46 & 0.6965 & 0.1846 \\
            Output-relative & 69.01 & 2.54$\times$ & 23.14 & 0.7669 & 0.1422 \\
            w/o Re-compute & 72.47 & 2.42$\times$ & 17.70 & 0.3984 & 0.4876 \\
            \rowcolor{linecolor}\textbf{Ours} & 69.11 & 2.54$\times$ & 24.74 & 0.8041 & 0.1121 \\
            \bottomrule
    \end{tabular}
\end{table}

\begin{table}[!t]
    \centering
    \footnotesize
    \caption{Ablation on update strategies for $k$.}
    \vspace{-7pt}
    \label{tab:ablation_k_update}
    \setlength{\tabcolsep}{1.5mm}
    \begin{tabular}{cccccc}
    \toprule
            \multirow{2.3}{*}{\makecell{Update \\ of $k$}} & \multicolumn{2}{c}{Efficiency} & \multicolumn{3}{c}{Visual Retention} \\ 
            \cmidrule(lr){2-3} \cmidrule(lr){4-6} 
            & Latency (s) $\downarrow$ & Speedup $\uparrow$ & PSNR $\uparrow$ & SSIM $\uparrow$ & LPIPS $\downarrow$ \\
            \midrule
            $k_{\text{avg}}$ & 107.40 & 1.63$\times$ & 27.18 & 0.8935 & 0.0624 \\
            $k_{\text{EMA}}$ & 73.14 & 2.40$\times$ & 24.94 & 0.8111 & 0.1073 \\
            \rowcolor{linecolor}$k_{\text{local}}$ & 69.11 & 2.54$\times$ & 24.74 & 0.8041 & 0.1121 \\
            \bottomrule
    \end{tabular}
\end{table}

\textbf{The effect of transformation rate $k$ update strategy.}
Accurate estimation of the transformation rate $k$ is essential for our effective adaptive caching. As shown in Tab.~\ref{tab:ablation_k_update}, we compare three update strategies: 1) history averaging ($k_{\text{avg}}$), which averages all $k$ values after the $R$ steps warm-up; 2) an EMA-like approach ($k_{\text{EMA}}$), utilizing an exponential moving average as $k_{\text{EMA}}^{(i)} = 0.9 k_{\text{EMA}}^{(i-1)} + 0.1 k^{(i)}$; and 3) our default local update ($k_{\text{local}}$), which uses the most recent fully computed interval from Eq.~\ref{eq:transformation-rate}. While $k_{\text{avg}}$ achieves the highest visual quality, it sacrifices speedup due to slow response to local dynamics. $k_{\text{EMA}}$ slightly improves PSNR over $k_{\text{local}}$, but at the cost of reduced efficiency. Our $k_{\text{local}}$ method achieves the best balance, providing the 2.54$\times$ speedup and competitive visual quality by promptly adapting to diffusion dynamics.

\begin{figure}[!t]
	\begin{center}
		\includegraphics[width=0.95\linewidth]{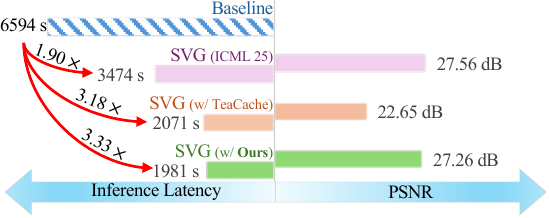}
	\end{center}
        \vspace{-5pt}
	\caption{The compatibility with other acceleration techniques (SVG)~\cite{xi2025sparse} on HunyuanVideo (129frames, 1280$\times$720).}
	\label{fig:compatibility_study}
\end{figure}

\begin{table}[!t]
    \centering
    \footnotesize
    \caption{Speedup comparison on GPU architectures.}
    \vspace{-5pt}
    \label{tab:speedup_gpu}
    \setlength{\tabcolsep}{4.mm}
    \begin{tabular}{lcc}
        \toprule
        Methods & Ampere \textnormal{\scriptsize{(A800)}} & Hopper \textnormal{\scriptsize{(H20)}} \\
        \midrule
        HunyuanVideo~\cite{kong2024hunyuanvideo} & 3064 s & 6594 s \\
        + SVG~\cite{xi2025sparse}   & 1959 s \textnormal{\scriptsize{(1.6$\times$)}} & 3474 s \textnormal{\scriptsize{(1.9$\times$)}} \\
        \rowcolor{linecolor}+ \textbf{Ours}  & \textbf{1335 s} \textnormal{\scriptsize{(\textbf{2.3$\times$})}} & \textbf{2897 s} \textnormal{\scriptsize{(\textbf{2.3$\times$})}} \\
        \bottomrule
    \end{tabular}
\end{table}

\subsection{Compatibility with Other Acceleration Techniques}

This subsection analyzes the flexibility of our approach by evaluating its compatibility with additional acceleration strategies on the time-consuming HunyuanVideo~\cite{kong2024hunyuanvideo}. Experiments are performed on 35 prompts uniformly sampled across seven dimensions of the T2V-CompBench~\cite{sun2024t2v} due to the high computational cost. As shown in Fig.~\ref{fig:compatibility_study}, generating a 5s 720P video takes 2 hours, even on an expensive H20 GPU. While the advanced, efficient attention method (SVG)~\cite{xi2025sparse} alone already achieves a 1.90$\times$ speedup with a PSNR of 27.56, integrating \ours~pushes the total speedup to an impressive 3.33$\times$, reducing the generation time for a high-resolution video from nearly 2 hours to just 33 minutes, with only a marginal 1.1\% PSNR drop. Although SVG combined with TeaCache~\cite{liu2025timestep} achieves a 3.18$\times$ speedup, it leads to a significant 17.8\% reduction in PSNR. We argue that TeaCache may fail to precisely identify the steps that are more important for visual retention. Notably, as shown in Tab.~\ref{tab:speedup_gpu}, our method results in consistent acceleration across both Hopper (H20) and Ampere (A800) architectures, whereas SVG’s gains are more hardware-dependent, making \ours~particularly suitable for diverse academic and practical environments.

\section{Conclusion}
This paper aims to alleviate the high computational cost of Diffusion Transformer-based video generation in a training-free manner. Our key insight is the relative stability of the transformation rate during iterative denoising, which enables dynamic computation reuse. Building on this, our \ours~employs a lightweight, runtime-adaptive criterion that leverages this intrinsic stability to reuse computations dynamically, eliminating the need for the costly offline profiling and external dataset priors that constrain existing methods. Extensive experiments show that \ours~achieves state-of-the-art acceleration while preserving visual fidelity. We believe our work offers a new perspective and a practical, efficient solution for accelerating diffusion models, facilitating their broader application.

\textbf{Limitation.} Although \ours~achieves an impressive 2.1–3.3$\times$ speedup, a significant gap remains from real-time generation. This is a common challenge mainly due to the need for large-scale models to generate high-quality videos. Enabling truly real-time video generation is an important direction for future research.

{\small
\bibliographystyle{IEEEtran}
\bibliography{references}
}

\end{document}